\documentclass[a4paper, 10pt, conference]{ieeeconf}      

\IEEEoverridecommandlockouts                              

\usepackage{graphics} 
\usepackage{epsfig} 
\usepackage{times} 
\usepackage{amsmath} 
\usepackage{cite}
 \usepackage{amsfonts}

\title{\LARGE \bf
Autonomous Driving at Intersections: \\A Critical-Turning-Point Approach for Left Turns
}

\author{Keqi Shu$^{1}$, Huilong Yu$^{1}$, Xingxin Chen$^{1}$, Long Chen$^{2}$, Qi Wang$^{3}$, Li Li$^{4}$, and Dongpu Cao$^{1}$
\thanks{$^{1}$K. Shu, H. Yu, X. Chen and D. Cao are with CogDrive lab, Department of Mechanical and Mechatronics Engineering, University of Waterloo, ON N2L 3G1, Canada {\tt\small k3shu, huilong.yu, xingxin.chen, dongpu.cao@uwaterloo.ca} (Corresponding authors: D. Cao and H. Yu)}%
\thanks{$^{2}$L. Chen is with the Waytous Inc., Qingdao 266109, China
        {\tt\small chen146@mail.sysu.edu.cn}}%
\thanks{$^{3}$Q. Wang is with the Autonomous Driving Lab, Tencent, China {\tt\small wangqi.dream@gmail.com}}%
\thanks{$^{4}$L. Li is  with the Tsinghua National Laboratory for Information Science and
Technology, Department of Automation, Tsinghua University, Beijing 100084,
China {\tt\small li-li@tsinghua.edu.cn}}%
}

\begin{document}

\setlength{\textfloatsep}{5pt}

\maketitle
\thispagestyle{empty}
\pagestyle{empty}

\begin{abstract}

Left-turn planning is one of the formidable challenges for autonomous vehicles, especially at unsignalized intersections due to the unknown intentions of oncoming vehicles. This paper addresses the challenge by proposing a \(critical\ turning\ point\) (\(CTP\)) based hierarchical planning approach. This includes a high-level candidate path generator and a low-level partially observable Markov decision process (POMDP) based planner. The proposed \(CTP\) concept, inspired by human-driving behaviors at intersections, aims to increase the computational efficiency of the low-level planner and to enable human-friendly autonomous driving. The POMDP based low-level planner takes unknown intentions of oncoming vehicles into considerations to perform less conservative yet safe actions. With proper integration, the proposed hierarchical approach is capable of achieving safe planning results with high commute efficiency at unsignalized intersections in real time.
\end{abstract}

\section{Introduction}

Unsignalized intersection is a difficult scenario for decision making and planning of autonomous vehicles (AVs) \cite{s0} due to the unknown intentions of surrounding human-driven vehicles. Left-turn planning at unsignalized intersections is one of the most common and dangerous tasks for autonomous vehicles, especially when the oncoming vehicle does not use turning signals. In such a complicated condition, balancing safety and commute efficiency is the key. Many approaches have been attempted to solve this contradiction.

Planning using predictions of surrounding vehicles is a popular approach \cite{s1,s2,s3}, which usually formulated the problem into a dynamic obstacle avoidance planning problem. This approach can be implemented in various scenarios with different prediction and planning algorithms. However, it is difficult to predict the accurate future path of each vehicle at intersections. The model predictive control approach \cite{s4,s5,s5plus} can solve planning and control problems simultaneously, but requires more precise vehicle and environment models, which affects real-time performance. 

The rule-based approach \cite{s6,s7,s7plus?} is easy to implement, but its performance is closely related to the modelling accuracy of the situations. For intersection planning problems, it would be nearly impossible to represent every scenario. Reinforcement learning \cite{s8,s9,s9plus} is another common approach for planning and decision making of AVs. However, the performance of the reinforcement learning approach is closely related to the construction of the Q-map, which needs to be large to have good performance.

Partially observable Markov decision process (POMDP) based approach \cite{s10,s10plus,s11} is an emerging technique for solving intersection planning problems in autonomous driving. With uncertain states being considered, the POMDP model could provide high efficiency at intersections. However, since solving a POMDP problem online using Monte Carlo sampling requires high computational power, actions are always set to be very discretized on a one-dimensional scale. Having two-dimensional action space (e.g. vertically and horizontally) enables exploration of the whole intersection space, but results in low-efficiency performance.

This paper proposes a novel approach to address left-turn planning problems at unsignalized intersections, where oncoming vehicles have unknown intentions. As for the original contributions, this paper:
\begin{itemize}
\item proposes a \(critical\ turning\ point\) (\(CTP\)) and POMDP based hierarchical planning framework;
\item proposes and validates a CTP concept for generating behavior-orientated paths;
\item formulates a two-dimensional left turn planning problem as a POMDP problem using the \(CTP\) concept, and solves the problem with good real-time performance.
\end{itemize}

The rest of the paper is structured as follows:  Section II introduces the novel \(CTP\) based hierarchical planning approach, while the POMDP problem formulation is presented in Section III. Section IV describes our simulation and results, and conclusions are presented in the last section.

\begin{figure*}
    \centering
    \includegraphics[width=0.9 \textwidth]{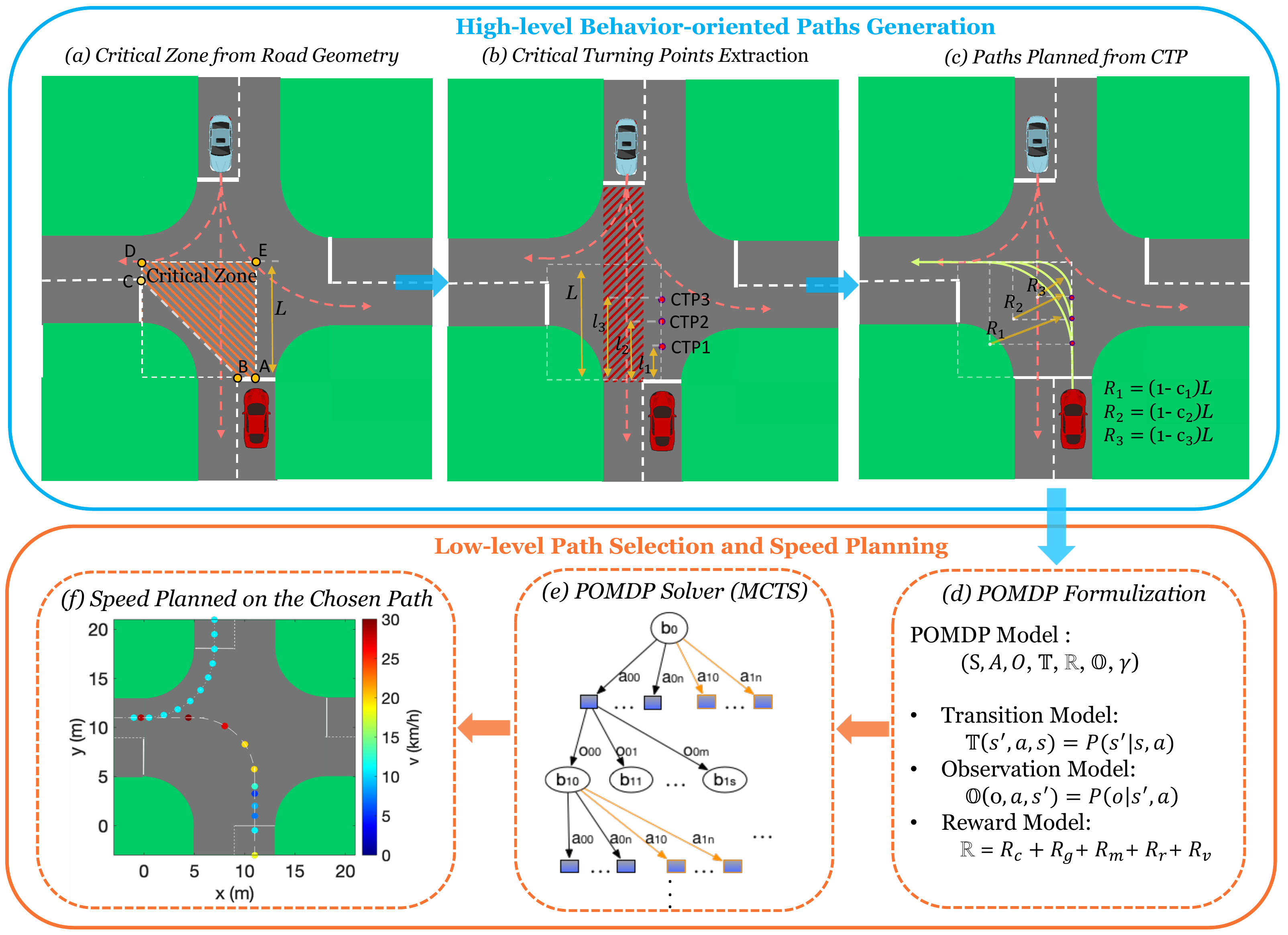}
    \caption{Configuration of the \(CTP\) based hierarchical planning approach for a standar intersection}
    \label{fig1}
\end{figure*}

\section{\(CTP\) based hierarchical planning approach}
The configuration of the \(CTP\) based hierarchical planning approach is demonstrated as Fig.~\ref{fig1}. The proposed framework can be divided into two parts: a high-level candidate path generation and a low-level path selection and speed planning. 

\subsection{High-level Behavior-oriented Paths Generation}
When making a left turn at an intersection with many oncoming vehicles, the left-turn vehicle usually creeps forward and waits at a certain position for the way to clear, then a sharp turn is performed from that point. Inspired by this phenomenon, a candidate path generation strategy is proposed for solving left-turn problems more efficiently, since the real-time performance is closely related to the proper modification of the POMDP model. The detailed procedure for high-level candidate paths generation will be explained in details below.

\subsubsection{\(Critical\ Zone\) Extraction From Road Geometry}
\(Critical\ Zone\) is the area that covers all the possible left-turn trajectories, which is presented as the orange shaded area in Fig.~\ref{fig1} (a). When the ego vehicle drives into an intersection, it uses the points A, B, C, D and E as shown in Fig.~\ref{fig1} (a) to generate the \(Critical\ Zone\). Positions of these points are closely related to the stop lines and road centerlines of the starting and target roads. Then the length (\(L\)) of the square-shape bounding box of \(critical\ zone\) will be sent to the next part of the proposed generator for \(CTP\)s extraction.

\subsubsection{\(CTP\) Extraction and Validation}
During left turns, there are some points where the ego vehicle shifts from creeping forward to make sharp steering and starts driving into the potential collision area
(red shaded area in Fig.~\ref{fig1}) (b). Decisions made around these points are crucial to guarantee safety, therefore, these points are defined as \(CTP\)s, and our proposed method defines the positions of each \(CTP\) as:
\begin{equation}\label{eq:1}
\begin{aligned}
{l_i = c_iL\ \ (i\in\{1,2,3\})}
\end{aligned}
\end{equation}
where \(l_1,\ l_2,\ l_3\) are the distances from the starting point of the left turn to the \(CTP\)s as shown in Fig.~\ref{fig1} (b). Each distance is calculated by multiplying the given length (\(L\)) of the \(critical\ zone\) with a certain ratio \(c_i\), this ratio is defined as \(critical\ ratio\). 

\begin{figure}
    \centering
    \includegraphics[width=0.45 \textwidth]{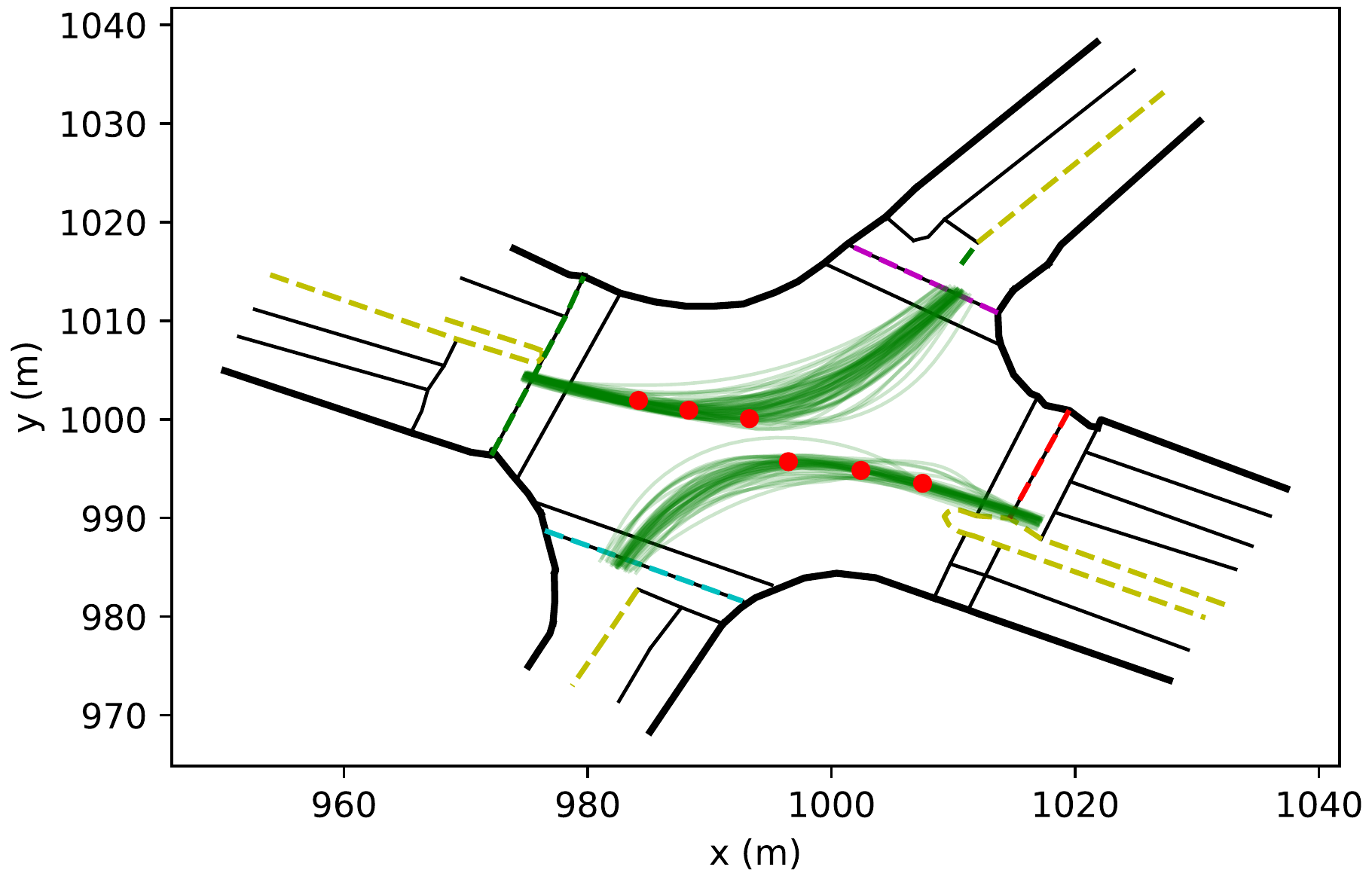}
    \caption{Representation of \(CTP\)s in real life}
    \label{fig3}
\end{figure}

To verify the rationality of the model, a recently released intersection traffic dataset \cite{m5} has been used. The dataset records the position, speed, steering angle, etc. of each vehicle at an intersection from a bird-eye view. Left-turn trajectories are extracted and plotted as the green lines in Fig.~\ref{fig3}. The starting points where the vehicle makes hard steering (sharp turning) are identified and extracted as turn points. The positions of these points are then clustered using the k-means algorithm, and the average position of each cluster is plotted as red dots in Fig.~\ref{fig3}, which proves that the aforementioned \(CTP\) does exist and also follows an isometric sequential pattern.

\subsubsection{Path Planned from \(CTP\)s and Coordinate Transfer}\label{AA}
With the starting point, \(CTP\)s and goal point as waypoints, candidate paths are generated using straight lines and quarter-circle curves. As shown in Fig.~\ref{fig1} (c), straight lines are used for paths planned from the starting point to \(CTP\)s. From the \(CTP\)s, quarter-circle curves are used for the ego vehicle to make the sharp turn until it faces to the left, then the rest of each path is filled with straight lines. The radii of each quarter-circle curve \(R_i\) as shown in Fig.~\ref{fig1} (c) are calculated using the length of the \(critical\ zone\) \(L\) in Fig.~\ref{fig1} (a) and the \(critical\ ratio\)s \(c_i\) in equation \ref{eq:1}: 
\begin{equation}
\begin{aligned}
{R_i = (1-c_i)L\quad (i\in\{1,2,3\})}
\end{aligned}
\end{equation}

After the candidate paths are generated in the Cartesian coordinate system, they are transferred into the Frenet frame \cite{m6} coordinate system. Since on a certain candidate path, the ego vehicle's travel distance \(s\) on the Frenet frame could be mapped to the Cartesian coordinate position \((x,y)\), the paths are reduced into a lower dimension, which are beneficial for searching efficiency for the model proposed later.

\subsection{Low-level Path Selection and Speed Planning}
Once candidate paths have been generated, they will be sent to the lower-level planner for path selection and speed planning. Proper POMDP modeling of the problem is made for the best representation of the scenarios, as shown in Fig.~\ref{fig1} (d). The detailed description of the POMDP model setup will be made in Section III. After the model has been properly set, as shown in Fig.~\ref{fig1} (e), a Monte Carlo tree search (MCTS) solver is used for optimal policy generation and action selection, which will be described in Section III.

\begin{figure*}
    \centering
    \includegraphics[width=0.9 \textwidth]{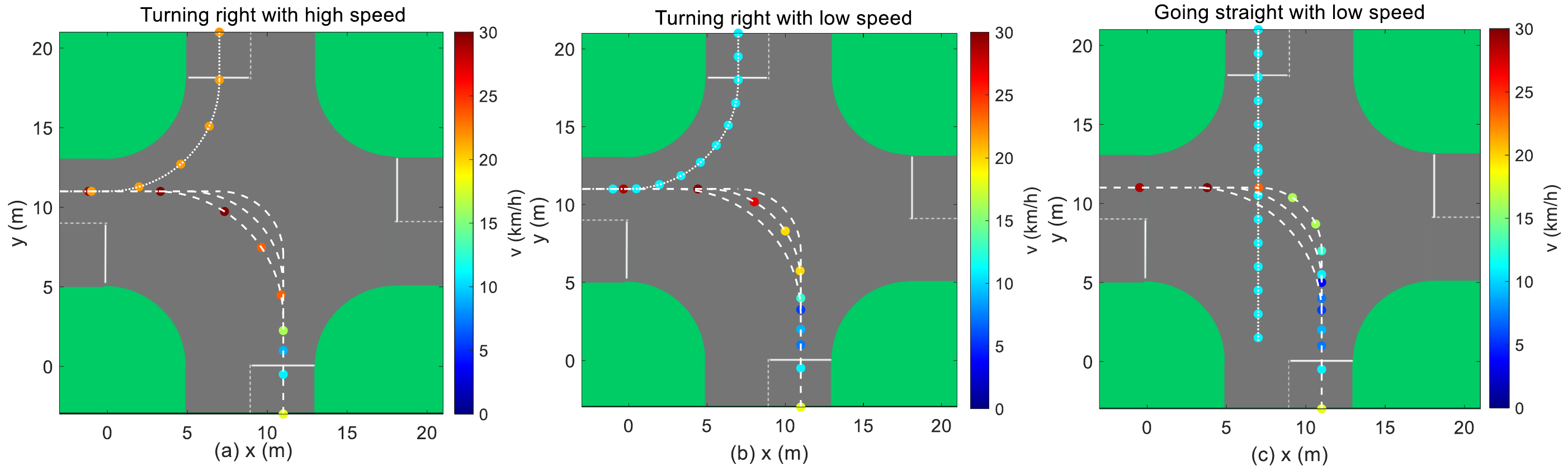}
    \caption{Simulation results of the proposed model}
    \label{fig5}
\end{figure*}

\section{POMDP problem formulation}\label{AA}
Partial observable Markov decision model is usually used for decision making in environments with uncertainties. The model could be formed into a tuple \((S,A,O,\mathbb{T},\mathbb{R},\mathbb{O},\gamma)\) as shown in Fig.~\ref{fig1} (d), where \(S,A,O\) are the state space, action space, observation space accordingly, and \(\mathbb{T},\mathbb{R},\mathbb{O}\) are the transition, reward, and observation model, while \(\gamma\) stands for the discount factor.

POMDP problems could be solved by off-line or on-line solvers. In our work, an online solver is chosen to adapt to the complexity of intersections. The Adaptive Belief Tree (ABT) \cite{m2} solver, which is an improved variation of the classical Monte Carlo tree search solver (POMCP) \cite{m1} is implemented. Instead of trimming the entire policy tree after an action is selected, the ABT algorithm only modifies parts of the tree that are influenced by the updated belief after executing the selected action. In this way, the search efficiency is boosted, which is critical for the safety and real-time efficiency at road intersection scenarios. 

\subsection{State Space and Observation Space}\label{AA}
The state space of the model is defined as the stacked states of all the vehicles, with the states considered as including both explicit and implicit states. Explicit states include directly observable information, and the implicit states are the intention of other drivers that is not directly observable but have influence on the ongoing states. With paths pre-planned in the previous section, Frenet frame is used to limit the dimensions of state space and action space. With the travelled distance and speed on a certain path, the global position and velocity of each vehicle can be defined with only two states instead of four.
The whole state space is written as
\begin{equation}
{X_t = (x_{0,t},x_{1,t}...,x_{n,t})} 
\end{equation}
with
\begin{equation}
{x_{i,t} = (S_{i,t},V_{i,t},P_{i,t}) \ \  i\in0,1,2,....,n}
\end{equation}
\(x_{0,t}\) represents the states of the ego vehicle at time \(t\). \(S_{0,t}\) stands for the travelled distance on the planned trajectory, \(V_{0,t}\) stands for the velocity along the trajectory and \(P_0\) is the path that is chosen by the ego vehicle. For states of the oncoming vehicle \( x_{i,t}\ (i=1,2...,n) \), the expanded state representation is similar, except \(P_{i,t}\) denotes left, right turn or straight heading for the oncoming vehicle.

For the observation model, it is assumed that there is no sensor noise in the scenario and the ego vehicle knows all its states. However, the future path of the oncoming vehicle is unknown. Thereby, the ego vehicle could only obtain the oncoming vehicle's global localization and speed in each time frame in the form of 
\begin{equation}
{O_t = (O_{1,t},O_{2,t},...,O_{n,t})}
\end{equation}
with each vehicle's observation at time \(t\), \(O_{n,t}\) can be written as \({(O_{n,t} = x_{n,t,} y_{n,t}, v_{n,t})}\). 

\subsection{Action sapce}\label{AA}
The action of the ego vehicle is defined in a compressed two-dimension space, including a acceleration variable \(a_v\) and a ”left-turn“ Boolean variable \(a_l\). The acceleration varies from -4 \(m/s^2\) to 4 \(m/s^2\) with a step of 1 \(m/s^2\). While the "left-turn" variable is Boolean (0 or 1) which conveys sharp turn instructions. 
\begin{equation}
{a = (a_v,a_l)}
\end{equation}

\subsection{Reward}\label{AA}
The reward function of the model is set as 
\begin{equation}
{R = R_c + R_g + R_v + R_m + R_r}
\end{equation}
where \(R_v = -(V - v_{ref})^2\) is set for the ego to follow the desired speed as close as possible on the pre-planned path, and the collision penalty \(R_c\) is defined as
\begin{equation}
  R_c =\begin{cases}
    -4000000, & \text{if $dist < dist_{safe}$}.\\
    0, & \text{otherwise}.
  \end{cases}
\end{equation}
similarly, the reward for reaching the goal is 
\begin{equation}
  R_g =\begin{cases}
    4000000, & \text{if $S > S_{goal}$}.\\
    0, & \text{otherwise}.
  \end{cases}
\end{equation}
and the penalty for moving backwards is defined as
\begin{equation}
  R_r =\begin{cases}
    -30000, & \text{if $V < 0$}.\\
    0, & \text{otherwise}.
  \end{cases}
\end{equation}
the marching reward \(R_m = 300y - 10x\) is set to encourage the ego vehicle to drive closer to the goal point on the \(x\), \(y\) axis. This is a crucial reward for the ego vehicle to select a proper path around \(CTP\)s during planning. 

\subsection{Transition Model}\label{AA}
Since all the vehicles are assumed to move on predefined routes, the transitions are defined on the Frenet frame as
\begin{equation}
{\boldsymbol{S_{t+1}} = \boldsymbol{S_t} + v_t\Delta t + 0.5 a_{acc} \Delta t^2 }\label{eq}
\end{equation}
\(a_{acc}\) stands for the acceleration of each vehicle, \(\boldsymbol{S_t}\) is the driving distance on the pre-planned path, and \(\Delta t\) is the time interval between each frame.

The oncoming vehicles are assumed to stay on the pre-determined route at a constant speed. 
As for the ego vehicle, if a sharper turn is processed, the ego vehicle's state variable \(P_e\) will be set to the number of the \(CTP\). If the ego vehicle passes the third \(CTP\), \(P_e\) is automatically set to three. After a \(CTP\) has been chosen, the ego vehicle only searches for optimal sequential actions on the designated path, and ignores all the sharp turn instruct actions.

\section{Simulations and results}
In the test scenario, there is one oncoming vehicle approaching the unsignalized intersection with a pre-defined intention (a route to follow) which is unknown to the ego vehicle. The ego vehicle is starting with the same position and speed in each scenario. Scenarios of different intentions and starting speeds of the oncoming vehicle have been tested. 

The planned trajectories and the speed of the ego and the oncoming vehicle in three particular scenarios are presented in Fig.~\ref{fig5}, the color of each dot corresponds to a certain speed. The three dashed lines are the candidate paths of the ego vehicle, and the dotted line is the pre-defined oncoming vehicle's path. In all of the scenarios in Fig.~\ref{fig5}, the ego vehicle pulls into the intersection with a relatively high speed, then it decelerates, maintains a low speed and performs the creeping forward behavior until the intention of the oncoming vehicle becomes certain. After the intention is clear, the rest of the path is planned with a definite assumption, and a more radical behavior is performed around the \(critical\ turning\ point\). Which shows the human-like characteristic of our approach and the rationality of our \(CTP\) model.

When the oncoming vehicle is going straight that blocks the left-turn paths as shown in Fig.~\ref{fig5} (c), the ego vehicle tends to take the candidate path with the furthest \(CTP\), and performs more radical actions around the third \(CTP\). This makes a lot of sense, since though taking a longer path is not an optimal option in a normal evaluation scheme, considering the moving nature of the oncoming vehicle, this decision decreases the time needed to travel through the intersection. Fig.~\ref{fig5} (a) and (b) shows that, in the scenarios that the oncoming vehicle is turning right, the ego vehicle intends to drive into the collision area at an early stage. This makes sense too, since if the right-turn intention of the oncoming vehicle is certain, the ego vehicle knows that the way will be quickly cleared for the left turn, which would be wise to make a sharper turn. It is also worth mentioning that, the ego vehicle picks different paths when the right-turn vehicle is driving with different speeds as shown in Fig.~\ref{fig5} (a) and (b). When the oncoming vehicle is driving with a lower speed (scenarios in Fig.~\ref{fig5} (b)), the waiting time before getting into the dangerous collision zone is longer, therefore the middle path is chosen. However, when the oncoming vehicle is driving with a higher speed (scenarios in Fig.~\ref{fig5} (a)), the shortest and the most aggressive path is chosen since the waiting time is shorter. Which shows the fully geometric exploration of our proposed method. 

\begin{figure}
    \centering
    \includegraphics[width=0.45 \textwidth]{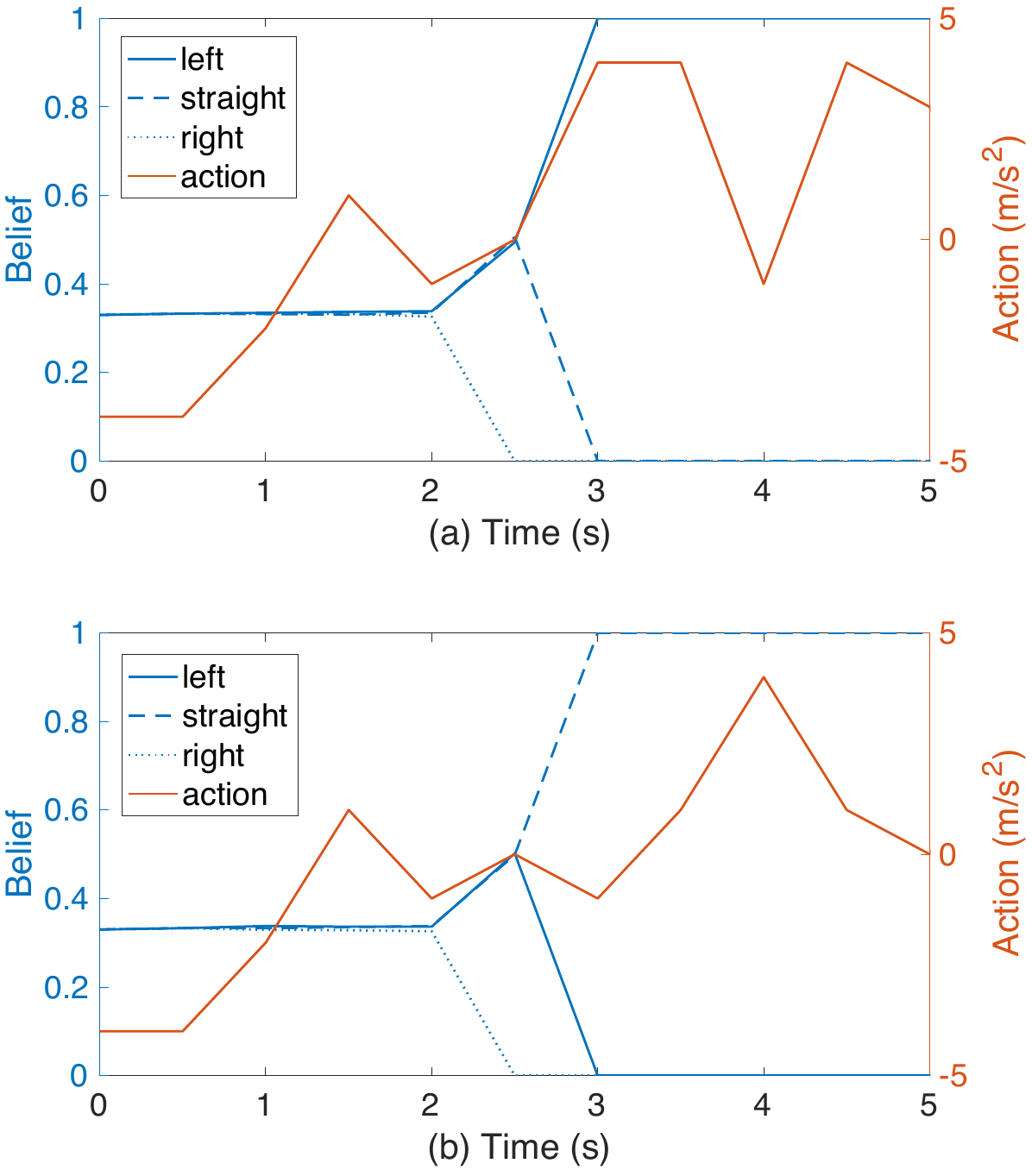}
    \caption{Belief-action correlated changing graph}
    \label{fig6}
\end{figure}

Fig.~\ref{fig6} shows how the actions of the ego vehicle change with changing beliefs of the intention of oncoming vehicles. As the ego vehicle travels forward, the intention of the oncoming vehicle becomes more and more certain. It can be seen from Fig.~\ref{fig6} (a) and (b) that before the intention of the oncoming vehicle is fully known, the actions share the same pattern. As shown in Fig.~\ref{fig6} (a), the moment when the oncoming traffic's left-turn intention becomes certain, an acceleration action is performed. Conversely, in Fig.~\ref{fig6} (b), when the oncoming vehicle's intention of going straight is assured, braking action is executed since the future path is blocked by the oncoming traffic.

\begin{figure}
    \centering
    \includegraphics[width=\columnwidth]{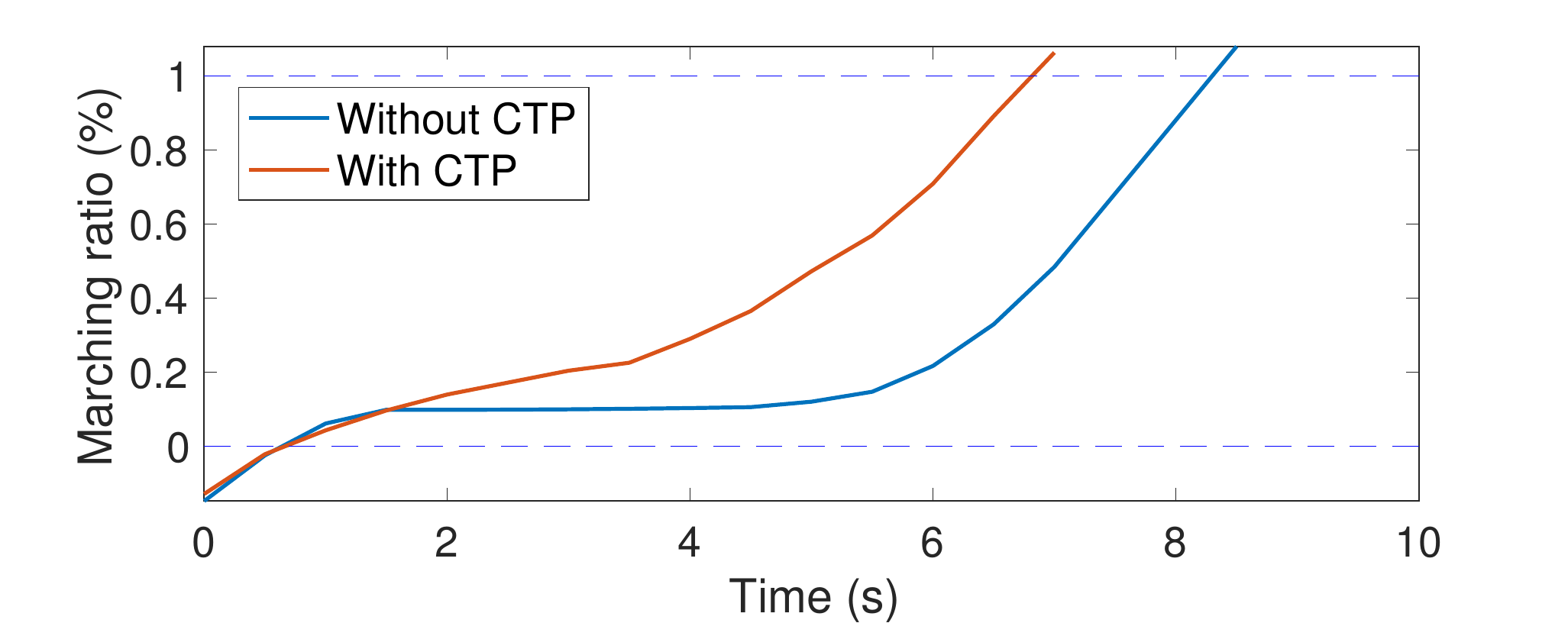}
    \caption{Marching ratio with respect to time}
    \label{fig7}
\end{figure}

The performance of commute efficiency at an intersection is also tested with our proposed planner. As shown in Fig.~\ref{fig7}, a comparison has been done between our planner (orange line in Fig.~\ref{fig7}) that generates paths with curves and straight lines using \(CTP\), and a planner (blue line in Fig.~\ref{fig7}) that only generates a path from road geometry (a quarter circle curve from the stop line to the goal point). The scenario is set with the oncoming vehicle driving straight at a constant speed while the ego vehicle tries to make a left turn. Since the length of paths that the two planners planned is different, a marching ratio is used to evaluate how far the ego vehicle has gone through. The marching ratio is calculated by taking the marching distance of the ego vehicle in the Frenet frame and divides it by the full length of the path planned by each vehicle. The dashed blue lines in Fig.~\ref{fig7} are the starting and ending marks of the intersection. The two curves both start with a negative marching ratio because the starting position of the ego vehicle is behind the stop line. Fig.~\ref{fig7} shows that our proposed method spends about 1.5s less time to pass through the intersection than the one that does not use \(CTP\)s. 

\begin{figure}
    \centering
    \includegraphics[width=\columnwidth]{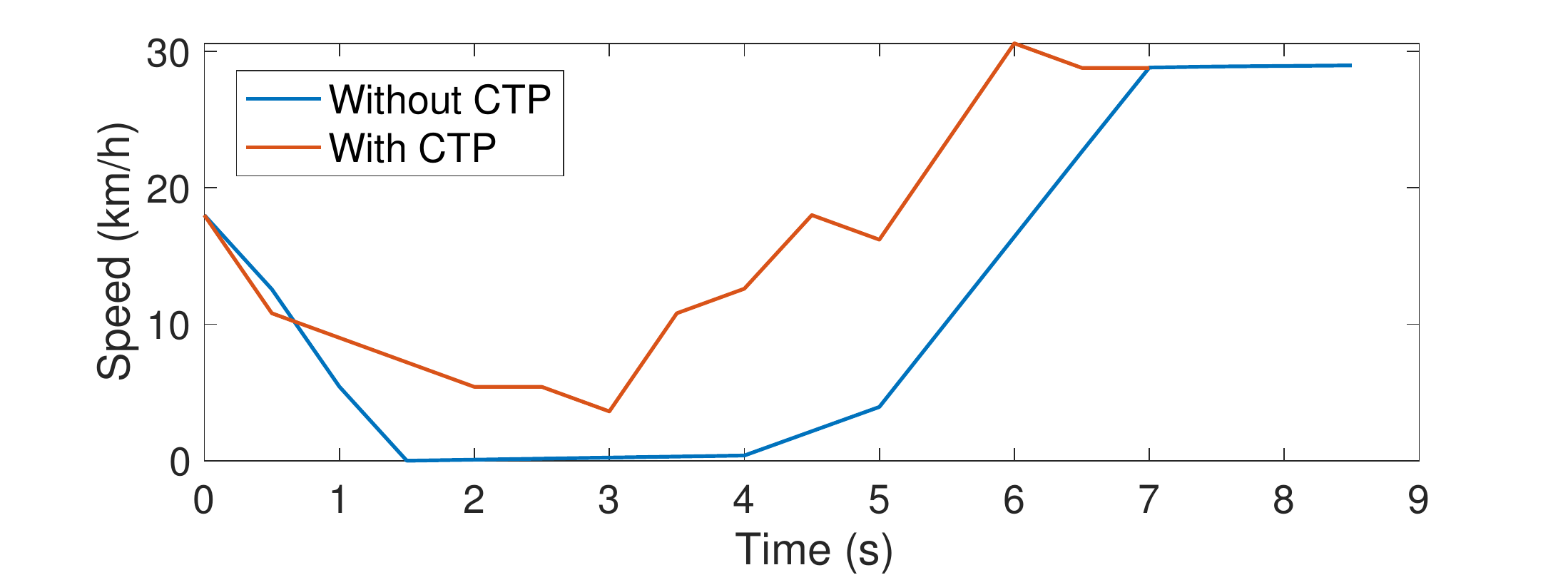}
    \caption{Speed profile on the left-turn path}
    \label{fig8}
\end{figure}

The reason that our proposed method has a higher efficiency in passing through the intersection can be explained by the result shown in Fig.~\ref{fig8}. Our candidate paths planned with \(CTP\) gives the ego vehicle an option to keep moving forward to reach the next \(CTP\) when one of the paths is blocked by the oncoming traffic. As shown in Fig.~\ref{fig8}, after 1.5 seconds, the oncoming traffics is blocking the way of the left-turn path. The planner that only uses road geometry as its path comes to a full stop, but with our proposed method, the ego vehicle keeps on creeping forward and waits closer to the goal point. When the path eventually becomes clear, the ego vehicle drives into the collision area with a shorter remaining distance. Since the ego vehicle keeps on creeping forward when the candidate paths are blocked, our proposed planner also obtains a higher average speed, which considerably increases the efficiency in passing the intersection.

\begin{figure}
    \centering
    \includegraphics[width=0.45\textwidth]{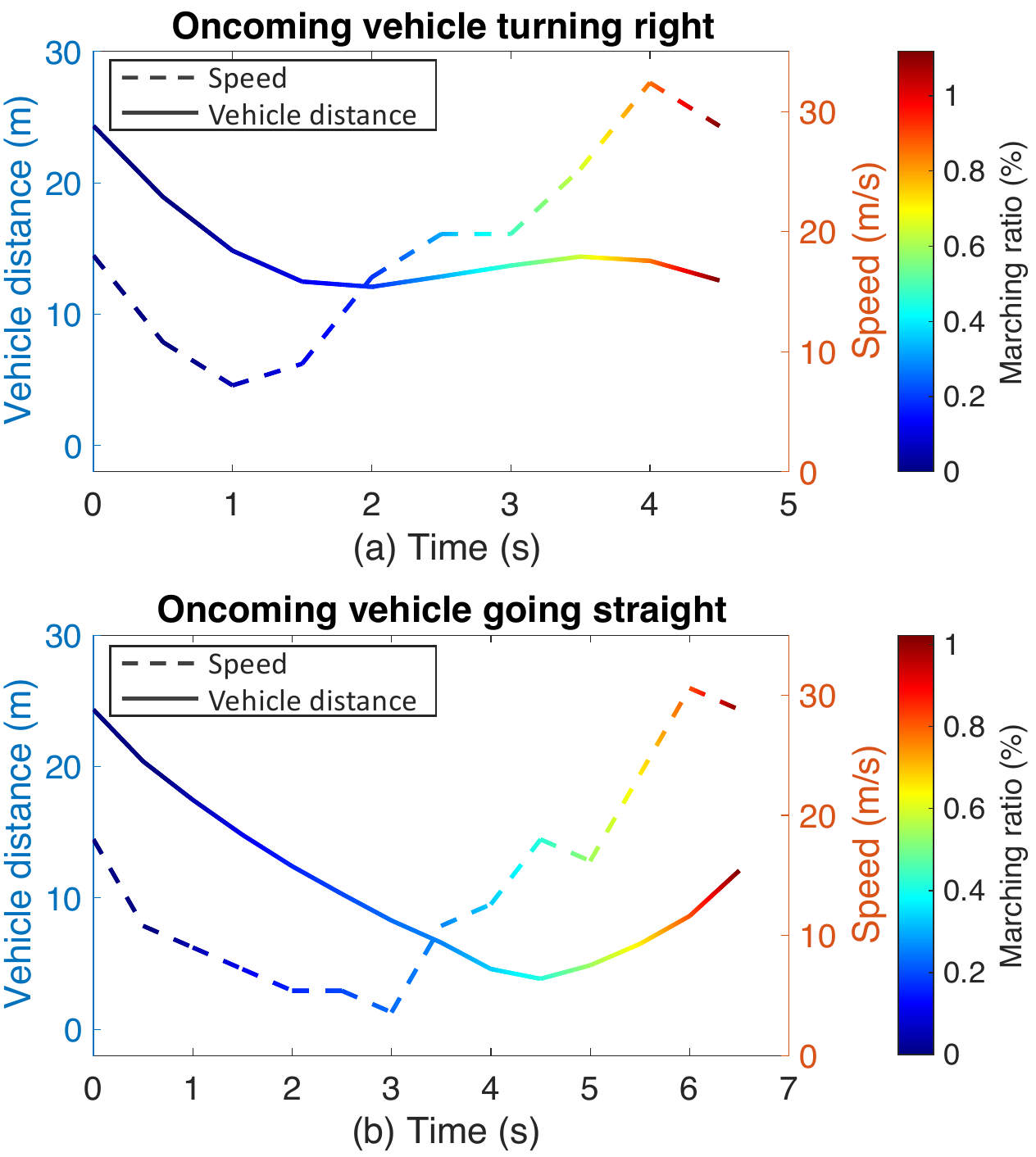}
    \caption{Vehicle distance, speed profile and marching ratio with respect to time in two critical scenarios}
    \label{fig9}
\end{figure}

Fig.~\ref{fig9} shows that our method is safety guaranteed. Two critical cases that the ego vehicle might crash with the oncoming traffic have been tested, which are the oncoming traffic is turning right and going straight. The solid lines are the distance between the ego vehicle and the oncoming vehicle. The dashed lines indicate the speed of the ego vehicle. The color bar indicates the marching ratio of the entire left turn process. It could be seen that the ego vehicle always keeps a safe distance to the oncoming vehicle and decreases speed before potential crashes occur. This indicates that our proposed method does not only guarantee safety in the current time frame, but also intends to avoid collisions in a future time frame.

In general, before merging into the intersection, the ego vehicle drives into the intersection with high speed. Then it decelerates to a lower speed and creeps forward, considering the potential of collision with the oncoming vehicle while waiting for the oncoming vehicle's intention to become certain. The ego vehicle then performs more confident actions on a proper route when the intention of the oncoming vehicles becomes clear. 

\section{Conclusions}
We have proposed a novel \(critical\ turning\ point\) (\(CTP\)) based hierarchical left-turn planning approach in this paper. This allows autonomous vehicles to perform safe and efficient planning behaviors as a driving expert during a left-turn scenario. We then proposed and validated a \(CTP\) concept for generating candidate paths for fast planning process and to enable friendly driving behavior of autonomous vehicles. A partially observable Markov decision process (POMDP) based solver is utilized to solve the formulated left-turn planning problems when taking into account the unknown intention of oncoming vehicles. With this proposed approach, urban driving of autonomous vehicles will be able to integrate into surrounding human-driven vehicles in a friendly manner, which as a result produces better commute efficiency. This is a high-level decision planner for general behavior planning. We will introduce a smoothing planer or controller in our future work.

\addtolength{\textheight}{-12cm}   







\bibliographystyle{unsrt}

\begin{thebibliography}{99}

\bibitem{s0} X. Li, Z. Sun, D. Cao, Z. He, and Q. Zhu, “Real-Time Trajectory Planning for Autonomous Urban Driving: Framework, Algorithms, and Verifications,” IEEE/ASME Transactions on Mechatronics, vol. 21, no. 2, pp. 740–753, 2016.
\bibitem{s1} S. Noh, “Decision-Making Framework for Autonomous Driving at Road Intersections: Safeguarding Against Collision, Overly Conservative Behavior, and Violation Vehicles,” IEEE Transactions on Industrial Electronics, vol. 66, no. 4, pp. 3275–3286, 2019.
\bibitem{s2} L. Ma, J. Xue, K. Kawabata, J. Zhu, C. Ma, and N. Zheng, “Efficient Sampling-Based Motion Planning for On-Road Autonomous Driving,” IEEE Transactions on Intelligent Transportation Systems, vol. 16, no. 4, pp. 1961–1976, 2015.
\bibitem{s3} C. Chen, M. Rickert, and A. Knoll, “Combining task and motion planning for intersection assistance systems,” 2016 IEEE Intelligent Vehicles Symposium (IV), 2016.
\bibitem{s4} G. Schildbach, M. Soppert, and F. Borrelli, “A collision avoidance system at intersections using Robust Model Predictive Control,” 2016 IEEE Intelligent Vehicles Symposium (IV), 2016.
\bibitem{s5} R. Hult, M. Zanon, S. Gros, and P. Falcone, “Optimal Coordination of Automated Vehicles at Intersections: Theory and Experiments,” IEEE Transactions on Control Systems Technology, vol. 27, no. 6, pp. 2510–2525, 2019.
\bibitem{s5plus} L. Huang and D. Panagou, “Automated turning and merging for autonomous vehicles using a Nonlinear Model Predictive Control approach,” 2017 American Control Conference (ACC), 2017.
\bibitem{s6} S. Klingelschmitt, F. Damerow, and J. Eggert, “Managing the complexity of inner-city scenes: An efficient situation hypotheses selection scheme,” 2015 IEEE Intelligent Vehicles Symposium (IV), 2015.
\bibitem{s7} R. Kohlhaas, T. Bittner, T. Schamm, and J. M. Zollner, “Semantic state space for high-level maneuver planning in structured traffic scenes,” 17th International IEEE Conference on Intelligent Transportation Systems (ITSC), 2014.
\bibitem{s7plus?} L. Zhao, R. Ichise, T. Yoshikawa, T. Naito, T. Kakinami, and Y. Sasaki, “Ontology-based decision making on uncontrolled intersections and narrow roads,” 2015 IEEE Intelligent Vehicles Symposium (IV), 2015.
\bibitem{s8} D. Isele, R. Rahimi, A. Cosgun, K. Subramanian, and K. Fujimura, “Navigating Occluded Intersections with Autonomous Vehicles Using Deep Reinforcement Learning,” 2018 IEEE International Conference on Robotics and Automation (ICRA), 2018.
\bibitem{s9} Z. Qiao, K. Muelling, J. M. Dolan, P. Palanisamy, and P. Mudalige, “Automatically Generated Curriculum based Reinforcement Learning for Autonomous Vehicles in Urban Environment,” 2018 IEEE Intelligent Vehicles Symposium (IV), 2018.
\bibitem{s9plus} D. Isele, R. Rahimi, A. Cosgun, K. Subramanian, and K. Fujimura, “Navigating Occluded Intersections with Autonomous Vehicles Using Deep Reinforcement Learning,” 2018 IEEE International Conference on Robotics and Automation (ICRA), 2018.
\bibitem{s10} M. Bouton, A. Cosgun, and M. J. Kochenderfer, “Belief state planning for autonomously navigating urban intersections,” 2017 IEEE Intelligent Vehicles Symposium (IV), 2017.
\bibitem{s10plus} C. Hubmann, J. Schulz, M. Becker, D. Althoff, and C. Stiller, “Automated Driving in Uncertain Environments: Planning With Interaction and Uncertain Maneuver Prediction,” IEEE Transactions on Intelligent Vehicles, vol. 3, no. 1, pp. 5–17, 2018.
\bibitem{s11} C. Hubmann, N. Quetschlich, J. Schulz, J. Bernhard, D. Althoff, and C. Stiller, “A POMDP Maneuver Planner For Occlusions in Urban Scenarios,” 2019 IEEE Intelligent Vehicles Symposium (IV), 2019.
\bibitem{s11plus} C.-W. Chang, C. Lv, H. Wang, H. Wang, D. Cao, E. Velenis, and F.-Y. Wang, “Multi-point turn decision making framework for human-like automated driving,” 2017 IEEE 20th International Conference on Intelligent Transportation Systems (ITSC), 2017.
\bibitem{m5} W. Zhan, L. Sun, \(et al\)., “INTERACTION Dataset: An INTERnational, Adversarial and Cooperative moTION Dataset in Interactive Driving Scenarios with Semantic Maps,” arXiv preprint arXiv:1910.03088, 2019.
\bibitem{m6} M. Werling, J. Ziegler, Kammel Sören, and S. Thrun, “Optimal trajectory generation for dynamic street scenarios in a Frenét Frame,” 2010 IEEE International Conference on Robotics and Automation, 2010.
\bibitem{m1} D. Silver and J. Veness, “Monte-Carlo planning in large POMDPs,” in Advances in Neural Information Processing Systems (NIPS), 2010.
\bibitem{m2} H. Kurniawati and V. Yadav, “An Online POMDP Solver for Uncertainty Planning in Dynamic Environment,” Springer Tracts in Advanced Robotics Robotics Research, pp. 611–629, 2016.
\end{thebibliography}

\end{document}